\documentclass[10pt,twocolumn,letterpaper]{article}

\usepackage{cvpr}
\usepackage{times}
\usepackage{epsfig}
\usepackage{graphicx}
\usepackage{amsmath}
\usepackage{amssymb}
\usepackage{enumitem}
\usepackage{booktabs}
\usepackage{multirow}
\usepackage{float}

\newcommand{\T}{\mathcal{T}}
\newcommand{\D}{\mathcal{D}}
\newcommand{\A}{\mathcal{A}}
\newcommand{\R}{\mathcal{R}}
\newcommand{\x}{\textbf{x}}

\DeclareMathOperator*{\argmin}{arg\,min}

\usepackage[pagebackref=true,breaklinks=true,letterpaper=true,colorlinks,bookmarks=false]{hyperref}

\cvprfinalcopy %

\def\cvprPaperID{4433} %

\ifcvprfinal\fi
\begin{document}

\title{Rethinking Few-Shot Image Classification: a Good Embedding Is All You Need?}

\author{
Yonglong Tian\textsuperscript{1*} \quad
Yue Wang\textsuperscript{1*} \quad
Dilip Krishnan\textsuperscript{2} \quad
Joshua B. Tenenbaum\textsuperscript{1} \quad
Phillip Isola\textsuperscript{1} \vspace{.3em}\\
\textsuperscript{1}MIT CSAIL \qquad\qquad \textsuperscript{2}Google Research
\vspace{-.5em}
}

\maketitle
\let\thefootnote\relax\footnotetext{*: equal contribution. 
}

\begin{abstract}
The focus of recent meta-learning research has been on the development of learning algorithms that can quickly adapt to test time tasks with limited data and low computational cost. Few-shot learning is widely used as one of the standard benchmarks in meta-learning. In this work, we show that a simple baseline: learning a supervised or self-supervised representation on the meta-training set, followed by training a linear classifier on top of this representation, outperforms state-of-the-art few-shot learning methods. An additional boost can be achieved through the use of self-distillation. This demonstrates that using a good learned embedding model can be more effective than sophisticated meta-learning algorithms. We believe that our findings motivate a rethinking of few-shot image classification benchmarks and the associated role of meta-learning algorithms. Code is available at:  \small{\url{http://github.com/WangYueFt/rfs/}}.
\end{abstract}
\section{Introduction}
Few-shot learning measures a model's ability to quickly adapt to new environments and tasks. This is a challenging problem because only limited data is available to adapt the model.
Recently, significant advances \cite{Wang-2016-4848,NIPS2016_6385,Triantafillou2017FewShotLT,pmlr-v70-finn17a,NIPS2017_6996,sung2018learning,Wang2018LowShotLF,NIPS2018_7352,rusu2018metalearning,YeHZS2018Learning,lee2019meta,li2019finding} have been made to tackle this problem using the ideas of meta-learning or ``learning to learn". %

Meta-learning defines a family of tasks, divided into disjoint meta-training and meta-testing sets. Each task consists of limited training data, which requires fast adaptability~\cite{NIPS2018_7293} of the learner (e.g., the deep network that is fine-tuned). During meta-training/testing, the learner is trained and evaluated on a task sampled from the task distribution. The performance of the learner is evaluated by the average test accuracy across many meta-testing tasks. Methods to tackle this problem can be cast into two main categories: optimization-based methods and metric-based methods. Optimization-based methods focus on designing algorithms that can quickly adapt to each task; while metric-based methods aim to find good metrics (usually kernel functions) to side-step the need for inner-loop optimization for each task. %

Meta-learning is evaluated on a number of domains such as few-shot classification and meta-reinforcement learning. Focusing on few-shot classification tasks, a question that has been raised in recent work is whether it is the meta-learning algorithm or the learned representation that is responsible for the fast adaption to test time tasks. \cite{raghu2019rapid} suggested that feature reuse is main factor for fast adaptation. Recently, \cite{Dhillon2019ABF} proposed transductive fine-tuning as a strong baseline for few-shot classification; and even in a regular, inductive, few-shot setup, they showed that fine-tuning is only slightly worse than state-of-the-art algorithms. In this setting, they fine-tuned the network on the meta-testing set and \emph{used} information from the testing data. Besides, \cite{chen19closerfewshot} shows an improved fine-tuning model performs slightly worse than meta-learning algorithms. 

In this paper, we propose an extremely simple baseline that suggests that good learned representations are more powerful for few-shot classification tasks than the current crop of complicated meta-learning algorithms. Our baseline consists of a \emph{linear} model learned on top of a pre-trained embedding. Surprisingly, we find this outperforms \emph{all other meta-learning algorithms} on few-shot classification tasks, often by large margins. The differences between our approach and that of \cite{Dhillon2019ABF} are: we \emph{do not} utilize information from testing data (since we believe that inductive learning is more generally applicable to few-shot learning); and we use a fixed neural network for feature extraction, rather than fine-tuning it on the meta-testing set. The findings in concurrent works~\cite{Chen2020ANM,Huang2019AllYN} are inline with our simple baseline. 

Our model learns representations by training a neural network on the entire meta-training set: we merge all meta-training data into a single task and a neural network is asked to perform either ordinary classification or self-supervised learning, on this combined dataset. The classification task is equivalent to the pre-training phase of TADAM \cite{NIPS2018_7352} and LEO \cite{rusu2018metalearning}. After training, we keep the pre-trained network up to the penultimate layer and use it as a feature extractor. During meta-testing, for each task, we fit a linear classifier on the features extracted by the pre-trained network. In contrast to \cite{Dhillon2019ABF} and  \cite{raghu2019rapid}, we \emph{do not} fine-tune the neural network.

Furthermore, we show that self-distillation on this baseline provides an additional boost. Self-distillation is a form of knowledge distillation \cite{knowledgedistillation}, where the student and teacher models are \emph{identical} in architecture and task. We apply self-distillation to the pre-trained network. %

\paragraph*{Contributions.} Our key contributions are:
\begin{itemize}[itemsep=0pt]
  \item A surprisingly simple baseline for few-shot learning, which achieves the state-of-the-art. This baseline suggests that many recent meta-learning algorithms are \emph{no better} than simply learning a good representation through a proxy task, e.g., image classification. 
  
  \item Building upon the simple baseline, we use self-distillation to further improve performance.
  
  \item Our combined method achieves an average of $3\%$ improvement over the previous state-of-the-art on widely used benchmarks. On the new benchmark Meta-Dataset~\cite{triantafillou2019meta}, our method outperforms previous best results by more than $7\%$ on average.
  
  \item Beyond supervised training, we show that representations learned with state-of-the-art self-supervised methods achieve similar performance as fully supervised methods. Thus we can ``learn to learn" simply by learning a good self-supervised embedding.%

\end{itemize}
\section{Related works}
\label{sec:related}

\paragraph{Metric-based meta-learning.} The core idea in metric-based meta-learning is related to nearest neighbor algorithms and kernel density estimation. Metric-based methods embed input data into fixed dimensional vectors and use them to design proper kernel functions. The predicted label of a query is the weighted sum of labels over support samples. Metric-based meta-learning aims to learn a task-dependent metric. \cite{Koch2015} used Siamese network to encode image pairs and predict confidence scores for each pair. Matching Networks \cite{NIPS2016_6385} employed two networks for query samples and support samples respectively and used an LSTM with read-attention to encode a full context embedding of support samples. Prototypical Networks \cite{NIPS2017_6996} learned to encode query samples and support samples into a shared embedding space; the metric used to classify query samples is the distance to prototype representations of each class. Instead of using distances of embeddings, Relation Networks \cite{sung2018learning} leveraged relational module to represent an appropriate metric. TADAM \cite{NIPS2018_7352} proposed metric scaling and metric task conditioning to boost the performance of Prototypical Networks.  

\paragraph{Optimization-based meta-learning.} Deep learning models are neither designed to train with very few examples nor to converge very fast. To fix that, optimization-based methods intend to learn with a few examples. Meta-learner \cite{ravi2017} exploited an LSTM to satisfy two main desiderata of few-shot learning: quick acquisition of task-dependent knowledge and slow extraction of transferable knowledge. MAML \cite{pmlr-v70-finn17a} proposed a general optimization algorithm; it aims to find a set of model parameters, such that a small number of gradient steps with a small amount of training data from a new task will produce large improvements on that task. In that paper, first-order MAML was also proposed, which ignored the second-order derivatives of MAML. It achieved comparable results to complete MAML with orders of magnitude speedup. To further simplify MAML, Reptile \cite{Nichol2018OnFM} removed re-initialization for each task, making it a more natural choice in certain settings. LEO \cite{rusu2018metalearning} proposed that it is beneficial to decouple the optimization-based meta-learning algorithms from high-dimensional model parameters. In particular, it learned a stochastic latent space from which the high-dimensional parameters can be generated. MetaOptNet \cite{lee2019meta} replaced the linear predictor with an SVM in the MAML framework; it incorporated a differentiable quadratic programming (QP) solver to allow end-to-end learning. For a complete list of recent works on meta-learning, we refer readers to \cite{weng2018metalearning}.

\paragraph{Towards understanding MAML.} To understand why MAML works in the first place, many efforts have been made either through an optimization perspective or a generalization perspective. Reptile \cite{Nichol2018OnFM} showed a variant of MAML works even without re-initialization for each task, because it tends to converge towards a solution that is close to each task's manifold of optimal solutions. In \cite{raghu2019rapid}, the authors analyzed whether the effectiveness of MAML is due to rapid learning of each task or reusing the high quality features. It concluded that feature reuse is the dominant component in MAML’s efficacy, which is reaffirmed by experiments conducted in this paper.

\paragraph{Meta-learning datasets.} Over the past several years, many datasets have been proposed to test meta-learning or few-shot learning algorithms. Omniglot~\cite{lake2015} was one of the earliest few-shot learning datasets; it contains thousands of handwritten characters from the world's alphabets, intended for one-shot "visual Turing test". In~\cite{Lake2019TheOC}, the authors reported the 3-year progress for the Omniglot challenge, concluding that human-level one-shot learnability is still hard for current meta-learning algorithms.  \cite{NIPS2016_6385} introduced mini-ImageNet, which is a subset of ImageNet~\cite{imagenet_cvpr09}. In~\cite{ren2018metalearning}, a large portion of ImageNet was used for few-shot learning tests. Meta-dataset~\cite{triantafillou2019meta} summarized recent datasets and tested several representative methods in a uniform fashion.

\paragraph{Knowledge distillation.} The idea of knowledge distillation (KD) dates back to \cite{Bucilua2006}. The original idea was to compress the knowledge contained in an ensemble of models into a single smaller model. In \cite{knowledgedistillation}, the authors generalized this idea and brought it into the deep learning framework. In KD, knowledge is transferred from the teacher model to the student model by minimizing a loss in which the target is the distribution of class probabilities induced by the teacher model. In was shown in \cite{yim2017} that KD has several benefits for optimization and knowledge transfer between tasks. BAN \cite{FurlanelloLTIA18} introduced sequential distillation, which also improved the performance of teacher models. In natural language processing (NLP), BAM \cite{clark2019bam} used BAN to distill from  single-task models to a multi-task model, helping the multi-task model surpass its single-task teachers. Another two related works are~\cite{mobahi2020self} which provides theoretical analysis of self-distillation and CRD~\cite{tian2020contrastive} which shows distillation improves the transferability across datasets. %

\section{Method}
We establish preliminaries about the meta-learning problem and related algorithms in \S\ref{sec:formulation}; then we present our baseline in \S\ref{sec:baseline}; finally, we introduce how knowledge distillation helps few-shot learning in \S\ref{sec:kd}. For ease of comparison to previous work, we use the same notation as \cite{lee2019meta}.

\subsection{Problem formulation}
\label{sec:formulation}
The collection of meta-training tasks is defined as $\T = \{(\D^{train}_i, \D^{test}_i)\}^I_{i=1}$, termed as meta-training set. The tuple $(\D^{train}_i, \D^{test}_i)$ describes a training and a testing dataset of a task, where each dataset contains a small number of examples. Training examples $\D^{train}=\{(\x_t, y_t)\}^T_{t=1}$ and testing examples $\D^{test}=\{(\x_q, y_q)\}^Q_{q=1}$ are sampled from the same distribution. 

A base learner $\A$, which is given by $y_* = f_\theta(\x_*)$ ($*$ denotes $t$ or $q$), is trained on $\D^{train}$ and used as a predictor on $\D^{test}$.  Due to the high dimensionality of $\x_*$, the base learner $\A$ suffers high variance. So training examples and testing examples are mapped into a feature space by an embedding model  $\boldsymbol{\Phi_*} = f_\phi(\x_*)$. Assume the embedding model is fixed during training the base learner on each task, then the objective of the base learner is
\begin{equation} \label{baselearner}
     \begin{split}
         \theta &= \A(\D^{train}; \phi) \\
               &=\argmin_\theta\mathcal{L}^{base}(\D^{train}; \theta, \phi) + \R(\theta),
     \end{split}
\end{equation}
where $\mathcal{L}$ is the loss function and $\R$ is the regularization term. 

The objective of the meta-learning algorithms is to learn a good embedding model, so that the average test error of the base learner on a distribution of tasks is minimized. Formally,
\begin{equation} \label{metalearner}
     \begin{split}
     \phi = \argmin_{\phi} \mathbb{E}_{\T}[\mathcal{L}^{meta}(\D^{test}; \theta, \phi)],
     \end{split}
\end{equation}
where $\theta= \A(\D^{train}; \phi)$.

Once meta-training is finished, the performance of the model is evaluated on a set of held-out tasks $\mathcal{S} = \{(\D^{train}_j, \D^{test}_j)\}^J_{j=1}$, called meta-testing set. The evaluation is done over the distribution of the test tasks:
\begin{equation} %
     \mathbb{E}_{\mathcal{S}}[\mathcal{L}^{meta}(\D^{test}; \theta, \phi), \text{where}~\theta= \A(\D^{train}; \phi)].
\end{equation}
\subsection{Learning embedding model through classification}
\label{sec:baseline}
\begin{figure}[t!]  
  \centering
 \includegraphics[width=1.0\columnwidth]{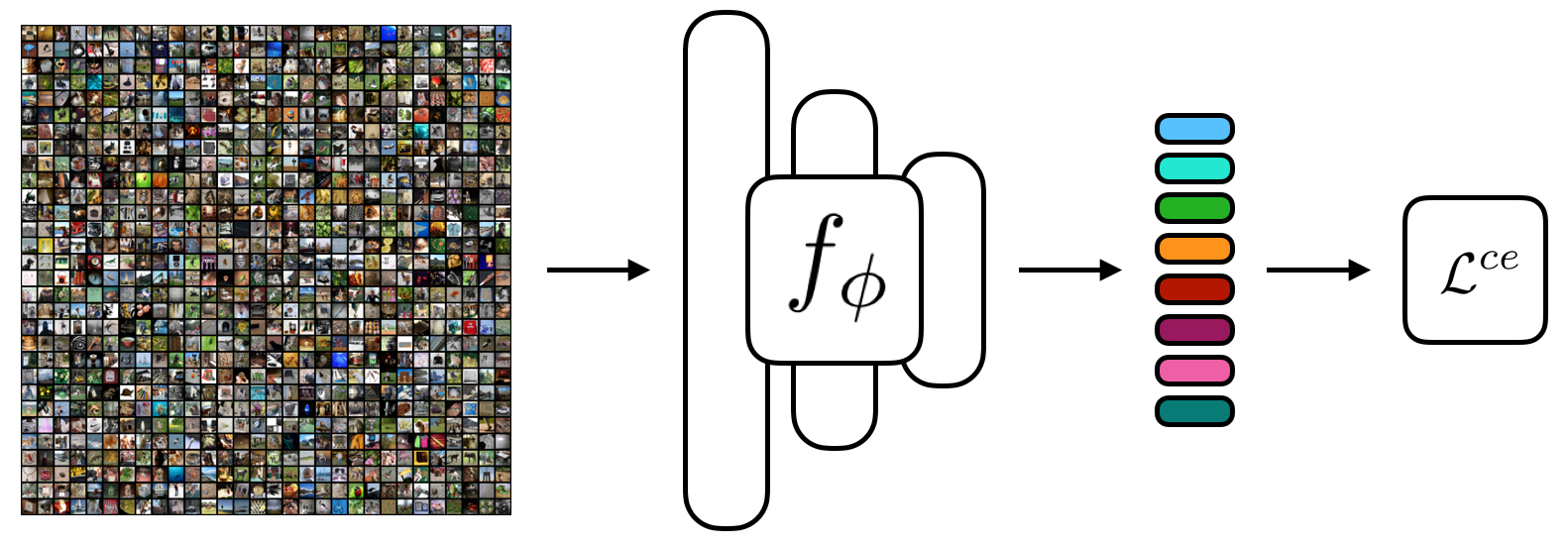}
  \caption{In meta-training, we train on an image classification task on the merged meta-training data to learn an embedding model. This model is then re-used at meta-testing time to extract embedding for a simple linear classifier.}
  \label{fig:meta-train}
\end{figure}

\begin{figure*}[t!]  
  \centering
 \includegraphics[width=1.6\columnwidth]{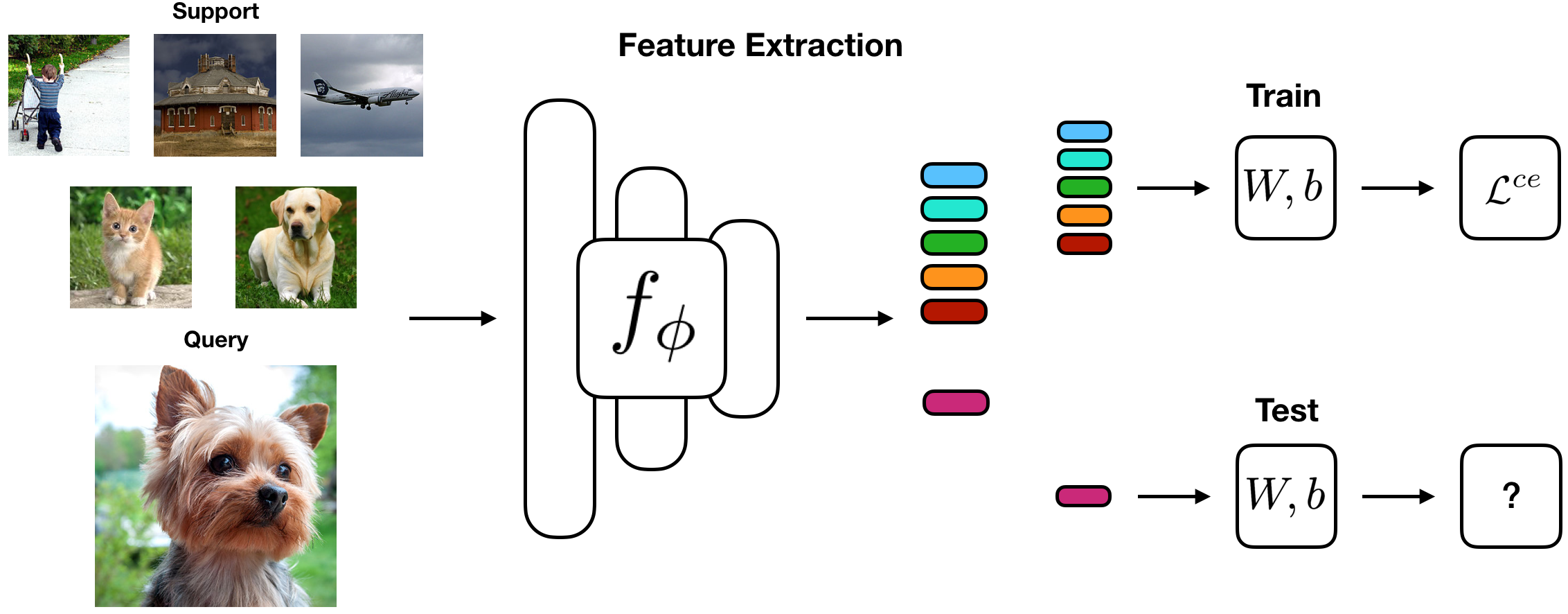}
  \caption{We show a meta-testing case for 5-way 1-shot task: 5 support images and 1 query image are transformed into embeddings using the fixed neural network; a linear model (logistic regression (LR) in this case) is trained on 5 support embeddings; the query image is tested using the linear model.}
  \label{fig:meta-test}
\end{figure*}

As we show in \S\ref{sec:formulation}, the goal of meta-training is to learn a transferrable embedding model $f_\phi$, which generalizes to any new task. Rather than designing new meta-learning algorithms to learn the embedding model, we propose that a model pre-trained on a classification task can generate powerful embeddings for the downstream base learner.  
To that end, we merge tasks from meta-training set into a single task, which is given by
\begin{equation} \label{mergetask}
     \begin{split}
        \D^{new} &= \{(\x_i, y_i)\}^K_{k=1} \\
                 &= \cup \{\D^{train}_1, \ldots, \D^{train}_i, \ldots, \D^{train}_I\},
     \end{split}
\end{equation}
where $\D^{train}_i$ is the task from $\T$. The embedding model is then
\begin{equation} \label{phi}
     \begin{split}
        \phi = \argmin_\phi \mathcal{L}^{ce} (\D^{new}; \phi),
     \end{split}
\end{equation}
and $\mathcal{L}^{ce}$ denotes the cross-entropy loss between predictions and ground-truth labels. We visualize the task in Figure~\ref{fig:meta-train}.

As shown in Figure~\ref{fig:meta-test}, for a task $(\D^{train}_j, \D^{test}_j)$ sampled from meta-testing distribution, we train a base learner on $\D^{train}_j$. The base learner is instantiated as multivariate logistic regression. Its parameters $\theta=\{\boldsymbol{W},\boldsymbol{b}\}$ include a weight term $\boldsymbol{W}$ and a bias term $\boldsymbol{b}$, given by
\begin{equation} \label{w}
     \begin{split}
        \theta = \argmin_{\{\boldsymbol{W},\boldsymbol{b}\}} \sum_{t=1}^T\mathcal{L}^{ce}_t (\boldsymbol{W}f_{\phi}(\x_t)+\boldsymbol{b}, y_t) + \R(\boldsymbol{W},\boldsymbol{b}).
     \end{split}
\end{equation}
We also evaluate other base learners such as nearest neighbor classifier with $\mathcal{L}$-2 distance and/or cosine distance in \S\ref{sec:choice-base-learner}. 

In our method, the crucial difference between meta-training and meta-testing is the embedding model parameterized by $\phi$ is carried over from meta-training to meta-testing and kept unchanged when evaluated on tasks sampled from meta-testing set. The base learner is re-initialized for every task and trained on $\D^{train}$ of meta-testing task. Our method is the same with the pre-training phase of methods used in \cite{rusu2018metalearning,NIPS2018_7352}. Unlike other methods \cite{Dhillon2019ABF,raghu2019rapid}, we \emph{do not} fine-tune the embedding model $f_\phi$ during the meta-testing stage. 

\subsection{Sequential self-distillation}
\label{sec:kd}
\begin{figure}[t!]  
  \centering
 \includegraphics[width=1.0\columnwidth]{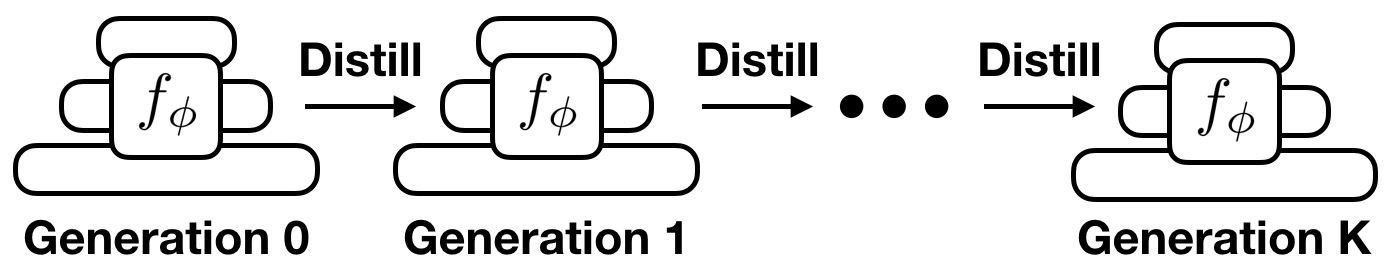}
  \caption{Sequential self-distillation: a vanilla model, termed as \emph{Generation 0}, is trained with standard cross-entropy loss; then, the $k$-th generation is learned with knowledge distilled from the ($k$-1)-th generation.}
  \label{fig:self-distill}
\end{figure}

Knowledge distillation \cite{knowledgedistillation} is an approach to transfer knowledge embedded in an ensemble of models to a single model, or from a larger teacher model to a smaller student model. Instead of using the embedding model directly for meta-testing, we distill the knowledge from the embedding model into a new model with an identical architecture, training on the same merged meta-training set. The new embedding model parameterized by $\phi^\prime$ is trained to minimize a weighted sum of the cross-entropy loss between the predictions and ground-truth labels and the Kullback–Leibler divergence (KL) between predictions and soft targets predicted by $f_{\phi}$:
\begin{equation} %
     \begin{split}
        \phi^\prime = \argmin_{\phi^\prime} & (\alpha \mathcal{L}^{ce} (\D^{new}; \phi^\prime) + \\ & \beta KL(f(\D^{new}; \phi^\prime), f(\D^{new};\phi))), 
     \end{split}
\end{equation}
where usually $\beta = 1-\alpha$. 

We exploit the Born-again~\cite{FurlanelloLTIA18} strategy to apply KD sequentially to generate multiple generations, which is shown in Figure~\ref{fig:self-distill}. At each step, the embedding model of $k$-th generation is trained with knowledge transferred from the embedding model of ($k$-1)-th generation:
\begin{equation} %
     \begin{split}
        \phi_k = \argmin_\phi & (\alpha \mathcal{L}^{ce} (\D^{new}; \phi) + \\ & \beta KL(f(\D^{new}; \phi), f(\D^{new};\phi_{k-1}))).
     \end{split}
\end{equation}
Assume we repeat the operation $K$ times, we use $\phi_K$ as the embedding model to extract features for meta-testing. We analyze the effects of sequential self-distillation in \S\ref{sec:exp_distill}. 
\section{Experiments}\label{sec:exp}
We conduct experiments on four widely used few-shot image recognition benchmarks: miniImageNet~\cite{NIPS2016_6385}, tieredImageNet~\cite{ren2018metalearning}, CIFAR-FS~\cite{bertinetto2018meta}, and FC100~\cite{NIPS2018_7352}. The first two are derivatives of ImageNet~\cite{ILSVRC15}, while the last two are reorganized from the standard CIFAR-100 dataset~\cite{Krizhevsky09learningmultiple,Torralba2008}. Additional results on Meta-Dataset~\cite{triantafillou2019meta} is presented in~\S\ref{sec:meta-dataset}.

\subsection{Setup}

\begin{table*}[tb]
\vspace{-10pt}
\begin{center}
\resizebox{0.95\linewidth}{!}{

\begin{small}
\begin{tabular}{@{}llc@{}cc@{}c@{}cc@{}}
\hline
\toprule
& & \phantom{a} & \multicolumn{2}{c}{\textbf{miniImageNet 5-way}} & \phantom{ab} & \multicolumn{2}{c}{\textbf{tieredImageNet 5-way}} \\
\cmidrule{4-5} \cmidrule{7-8}
\textbf{model} & \textbf{backbone} && \textbf{1-shot} & \textbf{5-shot} && \textbf{1-shot} & \textbf{5-shot}  \\
\midrule
MAML \cite{pmlr-v70-finn17a} & 32-32-32-32 &&  48.70 $\pm$ 1.84 & 63.11 $\pm$ 0.92 && 51.67 $\pm$ 1.81 & 70.30 $\pm$ 1.75 \\

Matching Networks \cite{NIPS2016_6385} & 64-64-64-64 && 43.56 $\pm$ 0.84 & 55.31 $\pm$ 0.73 && - & - \\

IMP \cite{pmlr-v97-allen19b} & 64-64-64-64 && 49.2 $\pm$ 0.7 & 64.7 $\pm$ 0.7 && - & - \\

Prototypical Networks\textsuperscript{$\dagger$} \cite{NIPS2017_6996} & 64-64-64-64 && 49.42 $\pm$ 0.78 & 68.20 $\pm$ 0.66 && 53.31 $\pm$ 0.89 & 72.69 $\pm$ 0.74 \\

TAML \cite{Jamal_2019_CVPR} & 64-64-64-64 && 51.77 $\pm$ 1.86 & 66.05 $\pm$ 0.85 && - & - \\

SAML \cite{Hao_2019_ICCV} & 64-64-64-64 && 52.22 $\pm$ n/a & 66.49 $\pm$ n/a && - & - \\

GCR \cite{Li_2019_ICCV} & 64-64-64-64 && 53.21 $\pm$ 0.80 & 72.34 $\pm$ 0.64 && - & - \\

KTN(Visual) \cite{Peng_2019_ICCV} & 64-64-64-64 && 54.61 $\pm$ 0.80 & 71.21 $\pm$ 0.66 && - & - \\

PARN\cite{Wu_2019_ICCV} & 64-64-64-64 && 55.22 $\pm$ 0.84 & 71.55 $\pm$ 0.66 && - & - \\

Dynamic Few-shot \cite{Gidaris_2018_CVPR} & 64-64-128-128 && 56.20 $\pm$ 0.86 & 73.00 $\pm$ 0.64 && - & - \\

Relation Networks \cite{sung2018learning} & 64-96-128-256 && 50.44 $\pm$ 0.82 & 65.32 $\pm$ 0.70 && 54.48 $\pm$ 0.93 & 71.32 $\pm$ 0.78 \\

R2D2~\cite{bertinetto2018meta} & 96-192-384-512 && 51.2 $\pm$ 0.6 & 68.8 $\pm$ 0.1 && - & -\\

SNAIL \cite{mishra2017simple} & ResNet-12 && 55.71 $\pm$ 0.99 & 68.88 $\pm$ 0.92 && - & - \\

AdaResNet \cite{munkhdalai2017rapid} & ResNet-12 && 56.88 $\pm$ 0.62 & 71.94 $\pm$ 0.57 && - & - \\

TADAM \cite{NIPS2018_7352} & ResNet-12 && 58.50 $\pm$ 0.30 & 76.70 $\pm$ 0.30 && - & - \\

Shot-Free~\cite{Ravichandran_2019_ICCV} & ResNet-12 && 59.04 $\pm$ n/a & 77.64 $\pm$ n/a && 63.52 $\pm$ n/a & 82.59 $\pm$ n/a \\

TEWAM~\cite{Qiao_2019_ICCV} & ResNet-12 && 60.07 $\pm$ n/a & 75.90 $\pm$ n/a && - & - \\

MTL~\cite{Sun_2019_CVPR} & ResNet-12 && 61.20 $\pm$ 1.80 & 75.50 $\pm$ 0.80 && - & - \\

Variational FSL~\cite{Zhang_2019_ICCV} & ResNet-12 && 61.23 $\pm$ 0.26 & 77.69 $\pm$ 0.17 && - & - \\

MetaOptNet~\cite{lee2019meta} & ResNet-12 && 62.64 $\pm$ 0.61 & 78.63 $\pm$ 0.46 && 65.99 $\pm$ 0.72 & 81.56 $\pm$ 0.53 \\

Diversity w/ Cooperation~\cite{Dvornik_2019_ICCV} & ResNet-18 && 59.48 $\pm$ 0.65 & 75.62 $\pm$ 0.48 && - & - \\

Fine-tuning~\cite{Dhillon2019ABF} & WRN-28-10 &&  57.73 $\pm$ 0.62 & 78.17 $\pm$ 0.49 && 66.58 $\pm$ 0.70 & 85.55 $\pm$ 0.48 \\

LEO-trainval\textsuperscript{$\dagger$} \cite{rusu2018metalearning} & WRN-28-10 && 61.76 $\pm$ 0.08 & 77.59 $\pm$ 0.12 && 66.33 $\pm$ 0.05 & 81.44 $\pm$ 0.09 \\

\midrule
Ours-simple & ResNet-12 && 62.02 $\pm$ 0.63 & 79.64 $\pm$ 0.44 && 69.74 $\pm$ 0.72 & 84.41 $\pm$ 0.55\\
Ours-distill & ResNet-12 && \textbf{64.82 $\pm$ 0.60} & \textbf{82.14 $\pm$ 0.43} && \textbf{71.52 $\pm$ 0.69} & \textbf{86.03 $\pm$ 0.49} \\
\bottomrule
\hline
\end{tabular}
\end{small}
}
\end{center}

\caption{
\textbf{Comparison to prior work on miniImageNet and tieredImageNet.} Average few-shot classification accuracies (\%) with 95\% confidence intervals on miniImageNet and tieredImageNet meta-test splits. Results reported with input image size of 84x84. a-b-c-d denotes a 4-layer convolutional network with a, b, c, and d filters in each layer. \textsuperscript{$\dagger$} results obtained by training on the union of training and validation sets.}
\vspace{-10pt}
\label{tab:miniImagenet}
\end{table*}

\textbf{Architecture.} Following previous works~\cite{mishra2017simple,NIPS2018_7352,lee2019meta,Ravichandran_2019_ICCV,Dhillon2019ABF}, we use a ResNet12 as our backbone: the network consists of 4 residual blocks, where each has 3 convolutional layers with 3$\times$3 kernel; a 2$\times$2 max-pooling layer is applied after each of the first 3 blocks; and a global average-pooling layer is on top of the fourth block to generate the feature embedding. Similar to ~\cite{lee2019meta}, we use Dropblock as a regularizer and change the number of filters from (64,128,256,512) to
(64,160,320,640). As a result, our ResNet12 is identical to that used in~\cite{Ravichandran_2019_ICCV,lee2019meta} .

\textbf{Optimization setup.} We use SGD optimizer with a momentum of 0.9 and a weight decay of $5e^{-4}$. Each batch consists of 64 samples. The learning rate is initialized as $0.05$ and decayed with a factor of $0.1$ by three times for all datasets, except for miniImageNet where we only decay twice as the third decay has no effect. We train 100 epochs for miniImageNet, 60 epochs for tieredImageNet, and 90 epochs for both CIFAR-FS and FC100. During distillation, we use the same learning schedule and set $\alpha=\beta=0.5$.

\textbf{Data augmentation.} When training the embedding network on transformed meta-training set, we adopt random crop, color jittering, and random horizontal flip as in \cite{lee2019meta}. 
For meta-testing stage, 
we train an $N$-way logistic regression base classifier. 
We use the implementations in scikit-learn \cite{sklearn} for the base classifier.

\subsection{Results on ImageNet derivatives}

\begin{table*}[t]

\vspace{-10pt}

\begin{center}
\resizebox{0.85\linewidth}{!}{
\begin{tabular}{@{}llc@{}cc@{}c@{}cc@{}}
\hline
\toprule
& & \phantom{a} & \multicolumn{2}{c}{\textbf{CIFAR-FS 5-way}} & \phantom{ab} & \multicolumn{2}{c}{\textbf{FC100 5-way}} \\
\cmidrule{4-5} \cmidrule{7-8}
\textbf{model} & \textbf{backbone} && \textbf{1-shot} & \textbf{5-shot} && \textbf{1-shot} & \textbf{5-shot}  \\

\midrule

MAML \cite{pmlr-v70-finn17a} & 32-32-32-32 &&  58.9 $\pm$ 1.9  & 71.5 $\pm$ 1.0  && - & - \\
Prototypical Networks \cite{NIPS2017_6996} & 64-64-64-64 && 55.5 $\pm$ 0.7 & 72.0 $\pm$ 0.6 && 35.3 $\pm$ 0.6 & 48.6 $\pm$ 0.6 \\
Relation Networks \cite{sung2018learning} & 64-96-128-256 && 55.0 $\pm$ 1.0 & 69.3 $\pm$ 0.8 && - & - \\
R2D2 \cite{bertinetto2018meta} & 96-192-384-512 && 65.3 $\pm$ 0.2 & 79.4 $\pm$ 0.1 && - & -\\
TADAM \cite{NIPS2018_7352} & ResNet-12 && - & - && 40.1 $\pm$ 0.4 & 56.1 $\pm$ 0.4\\

Shot-Free \cite{Ravichandran_2019_ICCV}& ResNet-12 && 69.2 $\pm$ n/a & 84.7 $\pm$ n/a && - & - \\

TEWAM~\cite{Qiao_2019_ICCV} & ResNet-12 && 70.4 $\pm$ n/a & 81.3 $\pm$ n/a && - & - \\

Prototypical Networks \cite{NIPS2017_6996}& ResNet-12 && 72.2 $\pm$ 0.7 & 83.5 $\pm$ 0.5 && 37.5 $\pm$ 0.6 & 52.5 $\pm$ 0.6 \\

MetaOptNet~\cite{lee2019meta} & ResNet-12 && 72.6 $\pm$ 0.7 & 84.3 $\pm$ 0.5 && 41.1 $\pm$ 0.6 & 55.5 $\pm$ 0.6 \\
\midrule
Ours-simple & ResNet-12 && 71.5 $\pm$ 0.8 & 86.0 $\pm$ 0.5 && 42.6 $\pm$ 0.7 & 59.1 $\pm$ 0.6 \\
Ours-distill & ResNet-12 && \textbf{73.9 $\pm$ 0.8} & \textbf{86.9 $\pm$ 0.5} && \textbf{44.6 $\pm$ 0.7} & \textbf{60.9 $\pm$ 0.6} \\

\bottomrule
\hline
\end{tabular}
}
\end{center}

\caption{\textbf{Comparison to prior work on CIFAR-FS and FC100.} Average few-shot classification accuracies (\%) with 95\% confidence intervals on CIFAR-FS and FC100. a-b-c-d denotes a 4-layer convolutional network with a, b, c, and d filters in each layer.} 
\label{tab:CIFAR}

\end{table*}

The miniImageNet dataset~\cite{NIPS2016_6385} is a standard benchmark for few-shot learning algorithms for recent works. It consists of 100 classes randomly sampled from the ImageNet; each class contains 600 downsampled images of size 84x84. We follow the widely-used splitting protocol proposed in~\cite{ravi2017}, which uses 64 classes for meta-training, 16 classes for meta-validation, and the remaining 20 classes for meta-testing.

The tieredImageNet dataset~\cite{ren2018metalearning} is another subset of ImageNet but has more classes (608 classes). These classes are first grouped into 34 higher-level categories, which are further divided into 20 training categories (351 classes), 6 validation categories (97 classes), and 8 testing categories (160 classes). Such construction ensures the training set is distinctive enough from the testing set and makes the problem more challenging.

\noindent\textbf{Results.} During meta-testing, we evaluate our method with 3 runs, where in each run the accuracy is the mean accuracy of $1000$ randomly sampled tasks. We report the median of 3 runs in Table~\ref{tab:miniImagenet}. Our simple baseline with ResNet-12 is already comparable with the state-of-the-art MetaOptNet~\cite{lee2019meta} on miniImageNet, and outperforms all previous works by at least 3\% on tieredImageNet. The network trained with distillation further improves over the simple baseline by 2-3\%.

We notice that previous works~\cite{Qiao_2018_CVPR,rusu2018metalearning,NIPS2018_7352,Sun_2019_CVPR} have also leveraged the standard cross-entropy pre-training on the meta-training set. In ~\cite{NIPS2018_7352,rusu2018metalearning}, a wide ResNet (WRN-28-10) is trained to classify all classes in the meta-training set (or combined meta-training and meta-validation set), and then frozen during the meta-training stage. \cite{Dhillon2019ABF} also conducts pre-training but the model is fine-tuned using the support images in meta-testing set, achieving $57.73 \pm 0.62$. We adopt the same architecture and gets $61.1 \pm 0.86$. So fine-tuning on small set of samples makes the performance worse. Another work~\cite{NIPS2018_7352} adopts a multi-task setting by jointly training on the standard classification task and few-shot classification (5-way) task. 
In another work~\cite{Sun_2019_CVPR}, the ResNet-12 is pre-trained before mining hard tasks for the meta-training stage. In this work, we show standard cross-entropy pre-training is sufficient to generate strong embeddings without meta-learning techniques or any fine-tuning.

\subsection{Results on CIFAR derivatives}

The CIFAR-FS dataset~\cite{bertinetto2018meta} is a derivative of the original CIFAR-100 dataset by randomly splitting 100 classes into 64, 16 and 20 classes for training, validation, and testing, respectively. The FC100 dataset~\cite{NIPS2018_7352} is also derived from  CIFAR-100 dataset in a similar way to tieredImagNnet.
This results in 60 classes for training, 20 classes for validation, and 20 classes for testing.

\noindent\textbf{Results.} Similar to previous experiments, we evaluate our method with 3 runs, where in each run the accuracy is the mean accuracy of 3000 randomly sampled tasks. Table~\ref{tab:CIFAR} summarizes the results, which shows that our simple baseline is comparable to Prototypical Networks ~\cite{NIPS2017_6996} and MetaOptNet~\cite{lee2019meta} on CIFAR-FS dataset, and outperforms both of them on FC100 dataset. Our distillation version achieves the new state-of-the-art on both datasets. This verifies our hypothesis that a good embedding plays an important role in few-shot recognition. 
\subsection{Embeddings from self-supervised representation learning}
\label{sec:self-sup}

\begin{table}[tb]
\begin{center}
\resizebox{1.0\linewidth}{!}{

\begin{small}
\begin{tabular}{@{}llc@{}cc@{}c@{}cc@{}}
\toprule
& & \phantom{a} & \multicolumn{2}{c}{\textbf{miniImageNet 5-way}}\\
\cmidrule{4-5} 
\textbf{model} & \textbf{backbone} && \textbf{1-shot} & \textbf{5-shot}  \\

\midrule
Supervised & ResNet50 && \textbf{57.56 $\pm$ 0.79} & 73.81 $\pm$ 0.63  \\
MoCo~\cite{He2019MomentumCF} & ResNet50 && 54.19 $\pm$ 0.93 & 73.04 $\pm$ 0.61  \\
CMC~\cite{tian2019contrastive} & ResNet50\textsuperscript{$\ast$} && 56.10 $\pm$ 0.89 & \textbf{73.87 $\pm$ 0.65}  \\

\bottomrule
\hline
\end{tabular}
\end{small}
}
\end{center}

\caption{
Comparsions of embeddings from supervised pre-training and self-supervised pre-training (Moco and CMC). \textsuperscript{$\ast$} the encoder of each view is 0.5$\times$ width of a normal ResNet-50.}
\label{tab:self-sup}
\end{table}

\begin{table*}[ht]
\centering
\small

\vspace{-10pt}
\resizebox{0.95\linewidth}{!}{

\setlength{\tabcolsep}{5.5pt}
\begin{tabular}{ ccccc|cc  cc  cc  cc}

\hline
\toprule

& & & &
& \multicolumn{2}{c}{\textbf{miniImageNet}}
& \multicolumn{2}{c}{\textbf{tieredImageNet}}
& \multicolumn{2}{c}{\textbf{CIFAR-FS}} 
& \multicolumn{2}{c}{\textbf{FC100}} \\

\textbf{NN} & \textbf{LR} & \textbf{$\mathcal{L}$-2} & \textbf{Aug} & \textbf{Distill}& 
1-shot & 5-shot &
1-shot & 5-shot &
1-shot & 5-shot &
1-shot & 5-shot \\

\midrule
\checkmark &  &  & &                         
 & 56.29 & 69.96 & 64.80 & 78.75 & 64.36 & 78.00  & 38.40 & 49.12 \\
 & \checkmark &  & &  
 & 58.74 & 78.31 & 67.62 & 84.77 & 66.92 & 84.78 & 40.36 & 57.23 \\
 & \checkmark & \checkmark &  &                  
 & 61.56 & 79.27 & 69.53 & 85.08 & 71.24 & 85.63 & 42.77 & 58.86 \\
 & \checkmark & \checkmark & \checkmark &           
 & 62.02 & 79.64 & 69.74 & 85.23 & 71.45 & 85.95 & 42.59 & 59.13 \\
 & \checkmark & \checkmark & \checkmark & \checkmark & 64.82 & 82.14 & 71.52 & 86.03 & 73.89 & 86.93 & 44.57 & 60.91\\

\bottomrule
\end{tabular}

}

\caption{\textbf{Ablation study on four benchmarks with ResNet-12 as backbone network.} ``NN'' and ``LR'' stand for nearest neighbour classifier and logistic regression. ``$\mathcal{L}$-2'' means feature normalization after which feature embeddings are on the unit sphere. ``Aug'' indicates that each support image is augmented into 5 samples to train the classifier. ``Distill'' represents the use of knowledge distillation. }
\label{table:ablation}
\end{table*}

\begin{figure*}[t]
\centering
\includegraphics[width=\linewidth]{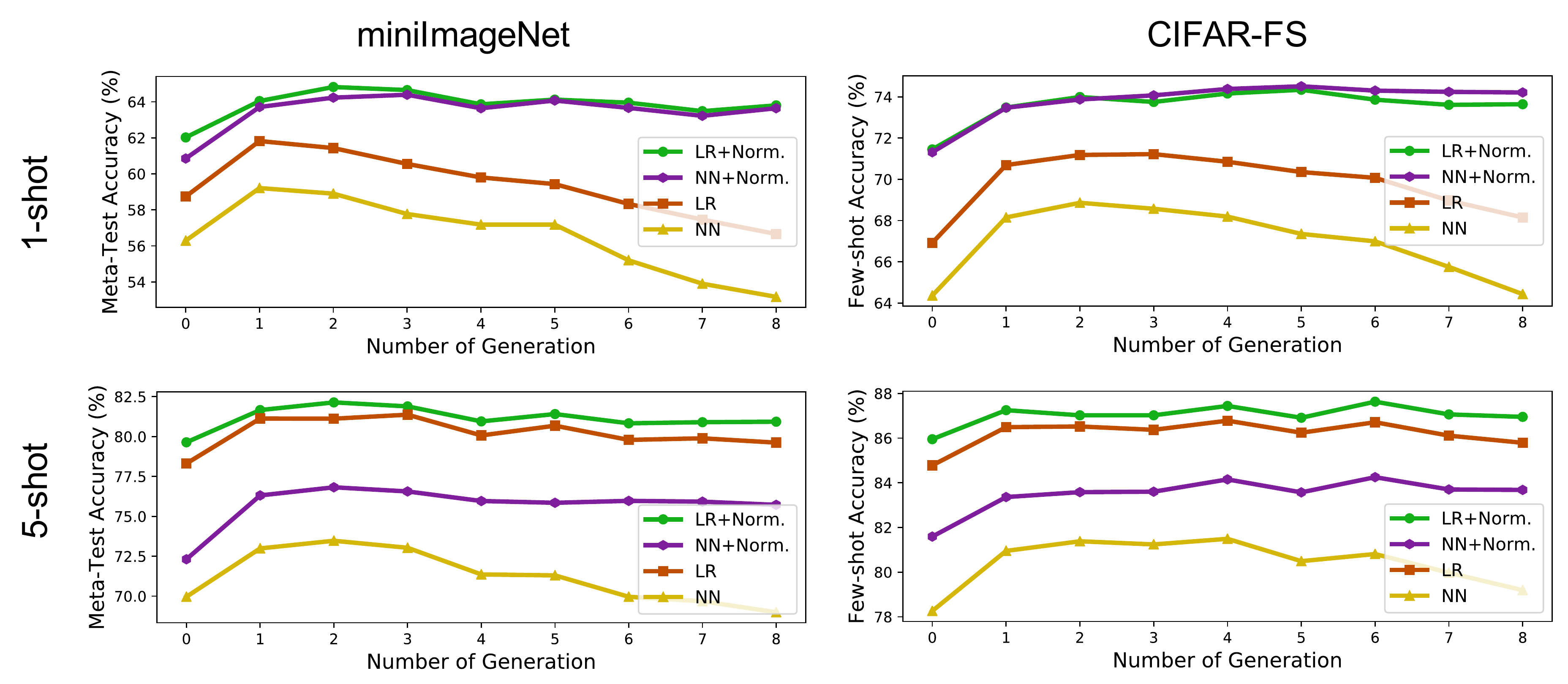}
\caption{\small{
\textbf{Evaluation on different generations of distilled networks.} 
The $0$-th generation (or root generation) indicates the vanilla network trained with only standard classification cross-entropy loss. The $k$-th generation is trained by combining the standard classification loss and the knowledge distillation (KD) loss using the ($k$-1)-th generation as the teacher model. Logistic regression (LR) and nearest neighbours (NN) are evaluated. 
}}
\label{fig:exp_distill}
\end{figure*}

Using unsupervised learning~\cite{wu2018unsupervised,tian2019contrastive,He2019MomentumCF,tian2020makes} to improve the generalization of the meta-learning algorithms~\cite{NIPS2016_6408} removes the needs of data annotation. In addition to using embeddings from supervised pre-training, we also train a linear classifier on embeddings from self-supervised representation learning. Following MoCo~\cite{He2019MomentumCF} and CMC~\cite{tian2019contrastive} (both are inspired by InstDis~\cite{wu2018unsupervised}), we train a ResNet50~\cite{He2015DeepRL} (without using labels) on the merged meta-training set to learn an embedding model. We compare unsupervised ResNet50 to a supervised ResNet50. 
From Table~\ref{tab:self-sup}, we observe that using embeddings from self-supervised ResNet50 is only slightly worse than using embeddings from supervised ResNet50 (in 5-shot setting, the results are comparable). This observation shows the potential of self-supervised learning in the scenario of few-shot learning.

\subsection{Ablation experiments}

In this section, we conduct ablation studies to analyze how each component affects the few-shot recognition performance. We study the following five components of our method: (a) we chose logistic regression as our base learner, and compare it to a nearest neighbour classifier with euclidean distance; (b) we find that normalizing the feature vectors onto the unit sphere, e.g., $\mathcal{L}$-2 normalization, could improve the classification of the downstream base classifier; (c) during meta-testing, we create 5 augmented samples from each support image to alleviate the data insufficiency problem, and using these augmented samples to train the linear classifier;
(d) we distill the embedding network on the training set by following the sequential distillation~\cite{FurlanelloLTIA18} strategy.

Table~\ref{table:ablation} shows the results of our ablation studies on miniImageNet, tieredImageNet, CIFAR-FS, and FC100. In general, logistic regression significantly outperforms the nearest neighbour classifier, especially for the 5-shot case; $\mathcal{L}$-2 normalization consistently improves the 1-shot accuracy by 2\% on all datasets; augmenting the support images leads to marginal improvement; even with all these techniques, distillation can still provide 2\% extra gain.%

\subsection{Effects of distillation}\label{sec:exp_distill}

\begin{table*}[tb]
\begin{center}
\resizebox{0.9\linewidth}{!}{

\setlength{\tabcolsep}{6pt}

\begin{small}
\begin{tabular}{@{}llc@{}cc@{}c@{}cc@{}}
\hline
\toprule
& & \phantom{a} & \multicolumn{2}{c}{\textbf{miniImageNet 5-way}} & \phantom{ab} & \multicolumn{2}{c}{\textbf{tieredImageNet 5-way}} \\
\cmidrule{4-5} \cmidrule{7-8}
\textbf{model} & \textbf{backbone} && \textbf{1-shot} & \textbf{5-shot} && \textbf{1-shot} & \textbf{5-shot}  \\

\midrule

Ours  & 64-64-64-64 && 55.25 $\pm$ 0.58 & 71.56 $\pm$ 0.52 && 56.18 $\pm$ 0.70 & 72.99 $\pm$ 0.55 \\

Ours-distill  & 64-64-64-64 && 55.88 $\pm$ 0.59 & 71.65 $\pm$ 0.51 && 56.76 $\pm$ 0.68 & 73.21 $\pm$ 0.54  \\

Ours-trainval  & 64-64-64-64 && 56.32 $\pm$ 0.58 & 72.46 $\pm$ 0.52 && 56.53 $\pm$ 0.68 & 73.15 $\pm$ 0.58  \\

Ours-distill-trainval  & 64-64-64-64 && \textbf{56.64 $\pm$ 0.58} & \textbf{72.85 $\pm$ 0.50} && \textbf{57.35 $\pm$ 0.70} & \textbf{73.98 $\pm$ 0.56} \\

\midrule

Ours  & ResNet-12 && 62.02 $\pm$ 0.63 & 79.64 $\pm$ 0.44 && 69.74 $\pm$ 0.72 & 84.41 $\pm$ 0.55 \\

Ours-distill  & ResNet-12 && 64.82 $\pm$ 0.60 & 82.14 $\pm$ 0.43 && 71.52 $\pm$ 0.69 & 86.03 $\pm$ 0.49 \\

Ours-trainval  & ResNet-12 && 63.59 $\pm$ 0.61 & 80.86 $\pm$ 0.47 && 71.12 $\pm$ 0.68 & 85.94 $\pm$ 0.46 \\

Ours-distill-trainval  & ResNet-12 && \textbf{66.58 $\pm$ 0.65} & \textbf{83.22 $\pm$ 0.39} && \textbf{72.98 $\pm$ 0.71} & \textbf{87.46 $\pm$ 0.44} \\

\midrule

Ours  & SEResNet-12 && 62.29 $\pm$ 0.60 & 79.94 $\pm$ 0.46 && 70.31 $\pm$ 0.70 & 85.22 $\pm$ 0.50 \\

Ours-distill  & SEResNet-12 && 65.96 $\pm$ 0.63 & 82.05 $\pm$ 0.46 && 71.72 $\pm$ 0.69 & 86.54 $\pm$ 0.49 \\

Ours-trainval  & SEResNet-12 && 64.07 $\pm$ 0.61 & 80.92 $\pm$ 0.43 && 71.76 $\pm$ 0.66 & 86.27 $\pm$ 0.45 \\

Ours-distill-trainval  & SEResNet-12 && \textbf{67.73 $\pm$ 0.63} & \textbf{83.35 $\pm$ 0.41} && \textbf{72.55 $\pm$ 0.69} & \textbf{86.72 $\pm$ 0.49} \\

\bottomrule
\end{tabular}
\end{small}
}
\end{center}

\caption{
Comparisons of different backbones on miniImageNet and tieredImageNet. }
\label{tab:backbone}
\end{table*}
\begin{table*}[tb]
\begin{center}

\resizebox{0.85\linewidth}{!}{

\setlength{\tabcolsep}{7pt}
\begin{small}

\begin{tabular}{@{}llc@{}cc@{}c@{}cc@{}}
\toprule
& & \phantom{a} & \multicolumn{2}{c}{\textbf{CIFAR-FS 5-way}} & \phantom{ab} & \multicolumn{2}{c}{\textbf{FC100 5-way}} \\
\cmidrule{4-5} \cmidrule{7-8}
\textbf{model} & \textbf{backbone} && \textbf{1-shot} & \textbf{5-shot} && \textbf{1-shot} & \textbf{5-shot}  \\

\midrule
Ours & 64-64-64-64 && 62.7 $\pm$ 0.8 & 78.7 $\pm$ 0.5 && 39.6 $\pm$ 0.6 & 53.5 $\pm$ 0.5 \\

Ours-distill & 64-64-64-64 && 63.8 $\pm$ 0.8 & 79.5 $\pm$ 0.5 && 40.3 $\pm$ 0.6 & 54.1 $\pm$ 0.5 \\

Ours-trainval & 64-64-64-64 && 63.5 $\pm$ 0.8 & 79.8 $\pm$ 0.5 && 43.2 $\pm$ 0.6 & 58.5 $\pm$ 0.5 \\
Ours-distill-trainval & 64-64-64-64 && \textbf{64.9 $\pm$ 0.8} & \textbf{80.3 $\pm$ 0.5} && \textbf{44.6 $\pm$ 0.6}& \textbf{59.2 $\pm$ 0.5} \\

\midrule
Ours & ResNet-12 && 71.5 $\pm$ 0.8 & 86.0 $\pm$ 0.5 && 42.6 $\pm$ 0.7 & 59.1 $\pm$ 0.6 \\
Ours-distill & ResNet-12 && 73.9 $\pm$ 0.8 & 86.9 $\pm$ 0.5 && 44.6 $\pm$ 0.7 & 60.9 $\pm$ 0.6 \\

Ours-trainval & ResNet-12 && 73.1 $\pm$ 0.8 & 86.7 $\pm$ 0.5 && 49.5 $\pm$ 0.7 & 66.4 $\pm$ 0.6 \\
Ours-distill-trainval & ResNet-12 && \textbf{75.4 $\pm$ 0.8} & \textbf{88.2 $\pm$ 0.5} && \textbf{51.6 $\pm$ 0.7}& \textbf{68.4 $\pm$ 0.6} \\

\midrule

Ours & SEResNet-12 && 72.0 $\pm$ 0.8 & 86.0 $\pm$ 0.6 && 43.4 $\pm$ 0.6 & 59.1 $\pm$ 0.6 \\
Ours-distill & SEResNet-12 && 74.2 $\pm$ 0.8 & 87.2 $\pm$ 0.5 && 44.9 $\pm$ 0.6 & 61.4 $\pm$ 0.6 \\

Ours-trainval & SEResNet-12 && 73.3 $\pm$ 0.8 & 86.8 $\pm$ 0.5 && 49.9 $\pm$ 0.7 & 66.8 $\pm$ 0.6 \\
Ours-distill-trainval & SEResNet-12 && \textbf{75.6 $\pm$ 0.8} & \textbf{88.2 $\pm$ 0.5} && \textbf{52.0 $\pm$ 0.7} & \textbf{68.8 $\pm$ 0.6} \\

\bottomrule
\end{tabular}
\end{small}
}
\end{center}
\caption{ Comparisons of different backbones on CIFAR-FS and FC100.}
\vspace{-5pt}
\label{tab:backbone_cifar}
\end{table*}

We can use sequential self-distillation to get an embedding model, similar to the one in Born-again networks~\cite{FurlanelloLTIA18}. We therefore investigate the effect of this strategy on the performance of downstream few-shot classification.

In addition to logistic regression and nearest-neighbour classifiers, we also look into a cosine similarity classifier, which is equivalent to the nearest-neighbour classifier but with normalized features (noted as ``NN+Norm.'').
The plots of 1-shot and 5-shot results on miniImageNet and CIFAR-FS are shown in Figure~\ref{fig:exp_distill}. The 0-th generation (or root generation) refers to the vanilla model trained with only standard cross-entropy loss, and the ($k$-$1$)-th generation is distilled into $k$-th generation. In general, few-shot recognition performance keeps getting better in the first two or three generations. After certain number of generations, the accuracy starts decreasing for logistic regression and nearest neighbour. Normalizing the features can significantly alleviate this problem.

In Table~\ref{tab:miniImagenet},  Table~\ref{tab:CIFAR}, and Table~\ref{table:ablation}, we evalute the model of the second generation on miniImageNet, CIFAR-FS and FC100 datasets; we use the first generation on tieredImageNet. Model selection is done on the validation set.

\subsection{Choice of base classifier}
\label{sec:choice-base-learner}
One might argue in the 1-shot case, that a linear classifier should behavior similarly to a nearest-neighbour classifier. However in Table~\ref{table:ablation} and Figure~\ref{fig:exp_distill}, we find that logistic regression is clearly better than nearest-neighbour. We argue that this is casued by the scale of the features. After we normalize the features by the $\mathcal{L}$-2 norm, logistic regression (``LR+Norm'') performs similarly to the nearest neighbour classifier (``NN+Norm.''), as shown in the first row of Figure~\ref{fig:exp_distill}. However, when increasing the size of the support set to 5, logistic regression is significantly better than nearest-neighbour even after feature normalization

\subsection{Comparsions of different network backbones.}

\begin{table*}[ht]
\begin{center}
\setlength{\tabcolsep}{12pt}
\begin{tabular}{l|c|cccc}
\toprule
\multicolumn{1}{c}{ } & \multicolumn{4}{c}{\textbf{Trained on ILSVRC train split}} \\
\midrule
\multirow{2}{*}{Dataset} & \multirow{2}{*}{Best from~\cite{triantafillou2019meta}} & LR & SVM & LR-distill & SVM-distill \\
&  & (ours) & (ours) & (ours) & (ours) \\
\midrule
ILSVRC        & 50.50 & 60.14 & 56.48 & \textbf{61.48} & 58.33\\
Omniglot      & 63.37 & 64.92 & 65.90 & 64.31 & \textbf{66.77}\\
Aircraft      & \textbf{68.69} & 63.12 & 61.43 & 62.32 & 64.23\\
Birds         & 68.66 & 77.69 & 74.61 & \textbf{79.47} & 76.63\\
Textures      & 69.05 & 78.59 & 74.25 & \textbf{79.28} & 76.66\\
Quick Draw    & 51.52 & \textbf{62.48} & 59.34 & 60.83 & 59.02\\
Fungi         & 39.96 & 47.12 & 41.76 & \textbf{48.53} & 44.51\\
VGG Flower    & 87.15 & \textbf{91.60} & 90.32 & 91.00 & 89.66\\
Traffic Signs & 66.79 & 77.51 & \textbf{78.94} & 76.33 & 78.64\\
MSCOCO        & 43.74 & 57.00 & 50.81 & \textbf{59.28} & 54.10\\
\midrule
Mean Accuracy & 60.94 & 68.02 & 65.38 & \textbf{68.28} & 66.86\\
\bottomrule
\end{tabular}
\end{center}
\caption{Results on Meta-Dataset. Average accuracy (\%) is reported with variable number of ways and shots, following the setup in~\cite{triantafillou2019meta}. We compare four variants of out method (LR, SVM, LR-distill, and SVM-distill) to the best accuracy over 7 methods in~\cite{triantafillou2019meta}. In each episode, 1000 tasks are sampled for evaluation.}
\label{tab:meta-dataset}
\end{table*}

Better backbone networks generally produce better results; this is also obvious in few-shot learning and/or meta-learning (as shown in Table~\ref{tab:miniImagenet}). To further verify our assumption that the key success of few-shot learning algorithms is due to the quality of embeddings, we compare three alternatives in Table~\ref{tab:backbone} and Table~\ref{tab:backbone_cifar}: a ConvNet with four four convolutional layers (64, 64, 64, 64); a ResNet12 as in Table~\ref{tab:miniImagenet}; a ResNet12 with sequeeze-and-excitation~\cite{hu2018squeeze} modules. For each model, we have four settings: training on meta-training set; training and distilling on meta-training set; training on meta-training set and meta-validation set; training and distilling on meta-training set and meta-validation set. The results consistently improve with more data and better networks. This is inline with our hypothesis: embeddings are the most critical factor to the performance of few-shot learning/meta learning algorithms;  better embeddings will lead to better few-shot testing performance (even with a simple linear classier). In addition, our ConvNet model also outperforms other few-shot learning and/or meta learning models using the same network. This verifies that in both small model regime (ConvNet) and large model regime (ResNet), few-shot learning and meta learning algorithms are \emph{no better} than learning a good embedding model.

\subsection{Multi-task vs multi-way classification?}
\label{sec:multitask}

We are interested in understanding whether the efficacy of our simple baseline is due to multi-task or multi-way classification. We compare to training an embedding model through \emph{multi-task} learning: a model with shared embedding network and different classification heads is constructed, where each head is only classifying the corresponding category; then we use the embedding model to extract features as we do with our baseline model. This achieves $58.53\pm 0.8$ on mini-ImageNet 5-way 1-shot case, compared to our baseline model which is $62.02\pm 0.63$. So we argue that the speciality of our setting, where the few-shot classification tasks are mutually exclusive and can be merged together into a single \emph{multi-way} classification task, makes the simple model effective.

\section{Results on Meta-Dataset}\label{sec:meta-dataset}

Meta-Dataset~\cite{triantafillou2019meta} is a new benchmark for evaluating few-shot methods in large-scale settings. Compared to miniImageNet and tieredImageNet, Meta-Dataset provides more diverse and realistic samples.  

\textbf{Setup.} 
The ILSVRC (ImageNet) subset consists of 712 classes for training, 158 classes for validation, and 130 classes for testing. We follow the setting in Meta-Dateset~\cite{triantafillou2019meta} where the embedding model is trained solely on the ILSVRC training split. We use ResNet-18~\cite{He2015DeepRL} as the backbone network. The input size is 128$\times$128. In the pre-training stage, we use SGD optimizer with a momentum of 0.9. The learning rate is initially 0.1 and decayed by a factor of 10 for every 30 epochs. We train the model for 90 epochs in total. The batch size is 256. We use standard data augmentation, including randomly resized crop and horizontal flip. In the distillation stage, we set $\alpha=0.5$ and $\beta=1.0$. We perform distillation twice and use the model from the second generation for meta-testing. We do not use test-time augmentation in meta-testing. 
In addition to logistic regression (LR), we also provide results of linear SVM for completeness. 

We select the best results from~\cite{triantafillou2019meta} for comparison -- for each testing subset, we pick the best accuracy over 7 methods and 3 different architectures including 4-layer ConvNet, Wide ResNet, and ResNet-18. As shown in Table~\ref{tab:meta-dataset}, our simple baselines clearly outperform the best results from~\cite{triantafillou2019meta} on 9 out of 10 testing datasets, often by a large margin. Our baseline method using LR outperforms previous best results by more than $7\%$ on average. Also, self-distillation improves \texttt{max(LR, SVM)} in 7 out of the 10 testing subsets. Moreover, we notice empirically that logistic regression (LR) performs better than linear SVM. 

\section{Discussion}
We have proposed a simple baseline for few-shot image classification in the meta-learning context. This approach has been underappreciated in the literature thus far. We show with numerous experiments that uch a simple baseline outperforms the current state-of-the-arts on four widely-used few-shot benchmarks. Combined with self-distillation, the performance further improves by 2-3\%. Even when meta-training labels are unavailable, it may be possible to leverage state of the art self-supervised learning approaches to learn very good embeddings for meta-testing tasks.

\noindent1. What is the intuition of this paper? \\
\textbf{A:} We hope this paper will shed new light on few-shot classification. We believe representations play an important role. Shown by our empirical experiments, a linear model can generalize well as long as a good representation of the data is given. 

\noindent2. Why does this simple baseline work? Is there anything that makes few-shot classification special? \\
\textbf{A:} Few-shot classification is a special case of meta-learning in terms of compositionality of tasks. Each task is an $K$-way classification problem, and on current benchmarks the classes, even between tasks, are all mutually exclusive. This means we can merge all $N$ of the $K$-way classification tasks into a single but harder $NK$-way classification task. Our finding is that training an embedding model on this new $NK$-way task turns out to transfer well to meta-testing set. On the other hand, we also find that self-supervised embedding, which does not explicitly require this $NK$ compositionality, achieves a similar level of performance. A concurrent work~\cite{Du2020FewShotLV} studies the representations for few-shot learning from the theoretical point of view. 

\noindent3. Does your work negate recent progress in meta-learning? \\
\textbf{A:} No. Meta-learning is much broader than just few-shot classification. Although we show a simple baseline outperforms other complicated meta-learning algorithms in few-shot classification, methods like MAML may still be favorable in other meta-learning domains (e.g., meta-reinforcement learning).%

\noindent4. Why does distillation work? What does it suggest? \\
\textbf{A:} The soft-labels \cite{knowledgedistillation} from the teacher model depict the fact that some classes are closer to each other than other classes. For example, a white goat is much more similar to a brown horse than to an airplane. But the one-hot label does not capture this. 
After being regularized by soft-labels, the network learns to capture the metric distance. From theoretical perspective, \cite{phuong2019towards} provides analysis for linear case. Ongoing work~\cite{Mobahi2020SelfDistillationAR} argues distillation amplifies regularization in Hilbert space.

{\small
\bibliographystyle{ieee_fullname}
\bibliography{main_cvpr}
}

\clearpage
\appendix

\section{Architectures}
\begin{figure}[ht]  
  \centering
 \includegraphics[width=0.95\columnwidth]{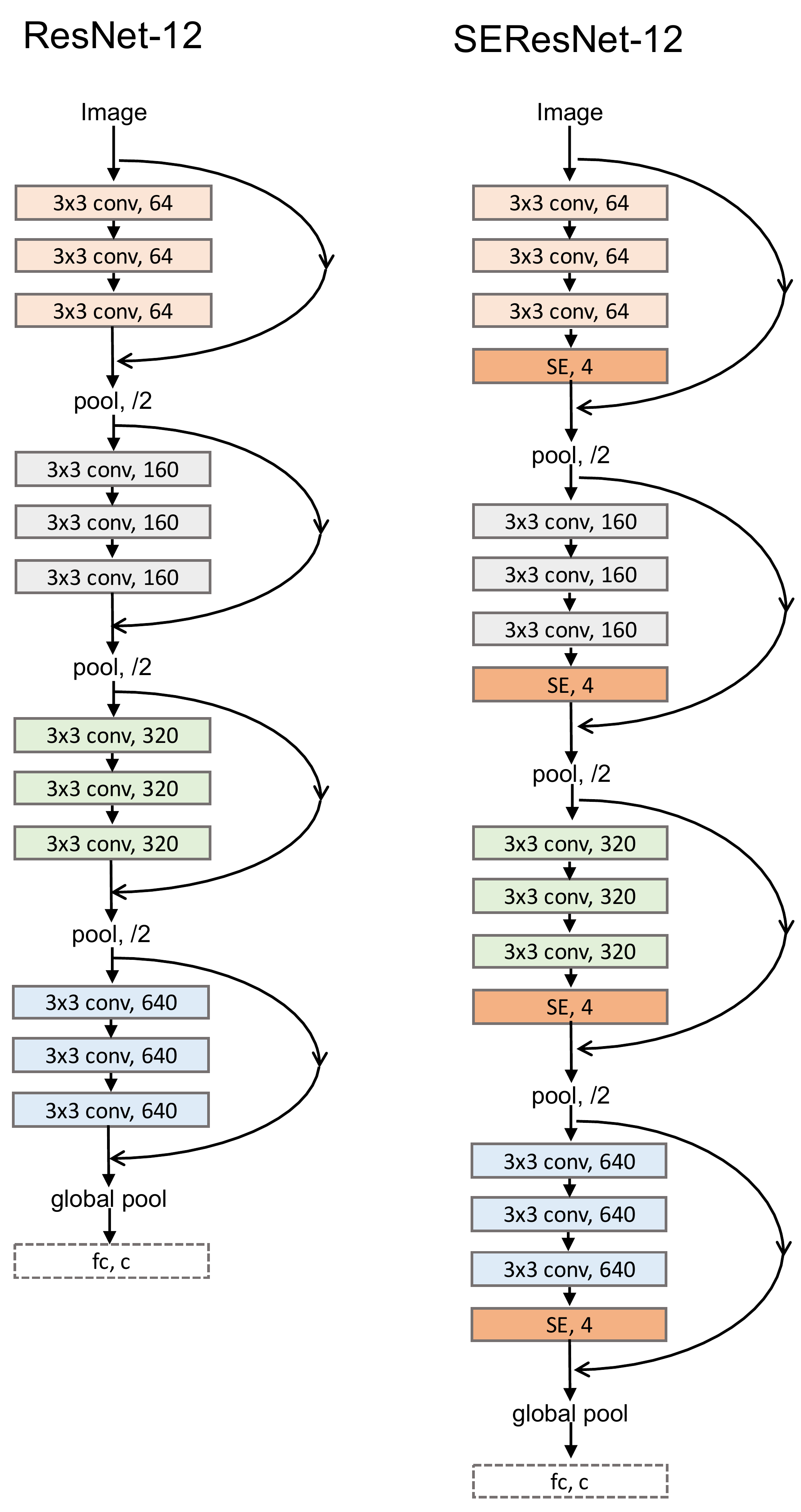}
  \caption{Nework architectures of ResNet-12 and SEResNet-12 used in this paper. The ``SE, 4'' stands for a Squeeze-and-Excitation layer with reduction parameter of 4. Dotted box will be removed during meta-testing stage.}
  \label{fig:arch}
\end{figure}
\noindent The architectures of ResNet-12 ans SEResNet-12 is show in Figure~\ref{fig:arch}. 

\section{More Training Details}
For SEResNet-12, we use the same training setup as ResNet-12 on all four benchmarks, as described in Sec 4.1. 

For 4-layer convnet, we also the same training setup as ResNet-12 on tieredImageNet, CIFAR-FS, and FC100, For miniImageNet, we train for 240 epochs with learning rate decayed at epochs 150, 180, and 210 with a factor of 0.1. We found that using the logit layer as feature results in slightly better accuracy ($\leq1\%$) on miniImageNet, so we report this number in Table~\ref{tab:backbone} for miniImageNet.

\section{Unsupervised Learning Details}
We adapt the first layer of a standard ResNet-50 to take images of size $84\times84$ as input. We only train on the meta-train set of miniImageNet dataset (do not use meta-val set). We follow the training recipe in CMC~\cite{tian2019contrastive} and MoCo~\cite{He2019MomentumCF} (which also follows InstDis~\cite{wu2018unsupervised}) except for two differences. The first one is that we only use $2048$ negatives for each positive sample as miniImageNet contains less than $40$k images in total. The second difference is that we train for $2000$ epochs, with a learning rate initialized as $0.03$ and decayed by consine annealing.

\end{document}